# JOINT EXTRACTION OF ENTITY AND RELATION WITH INFORMATION REDUNDANCY ELIMINATION

Yuanhao Shen and Jungang Han

School of computer science and Technology, Xi`an University of Posts and Telecommunications, Xi`an, China

***ABSTRACT***

*To solve the problem of redundant information and overlapping relations of the entity and relation extraction model, we propose a joint extraction model. This model can directly extract multiple pairs of related entities without generating unrelated redundant information. We also propose a recurrent neural network named Encoder-LSTM that enhances the ability of recurrent units to model sentences. Specifically, the joint model includes three sub-modules: the Named Entity Recognition sub-module consisted of a pre-trained language model and an LSTM decoder layer, the Entity Pair Extraction sub-module which uses Encoder-LSTM network to model the order relationship between related entity pairs, and the Relation Classification sub-module including Attention mechanism. We conducted experiments on the public datasets ADE and CoNLL04 to evaluate the effectiveness of our model. The results show that the proposed model achieves good performance in the task of entity and relation extraction and can greatly reduce the amount of redundant information.*

## 1. INTRODUCTION

Extraction of entity and relation, a core task in the field of Natural Language Processing (NLP), can automatically extract the entities and their relations from unstructured text. The results of this task play a vital role in various advanced NLP applications, such as knowledge map construction, question answering, and machine translation.

Supervised extraction of entity and relation usually uses a pipelined or joint learning approach. The pipelined approach treats the extraction task as two serial sub-tasks: named entity recognition [1] and relation classification. The relation classification sub-task first pairs the identified entities according to some pairing strategy, and then classifies the relationships between the entities. Due to the small number of entities that are related, the pipelined model usually generates a large number of pairs of unrelated entities during the pairing phase. Besides, the method also suffered from error propagating and paying little attention to the relevance of the two sub-tasks. To tackle the problems, researchers have conducted a lot of research on the joint learning and achieved better results. Joint Learning refers to extracting entities and classifying relations by one joint model. The joint models usually adopt three research ideas: parameter sharing [2], [3], [4], multi-head selection [5], [6], [7], and table filling [8], [9], [10]. These ideas take advantage of the relevance of sub-tasks to mitigate the error propagation, but still have to deal with the redundant information of unrelated entity pairs. Eberts *et al.* [11] proposed a span-based joint model that relies on the pre-trained Transformer network BERT as its core. The model achieved excellent performance but still suffered from the redundancy problem. Zheng *et al.* [12] proposed a method that uses a novel labeling mechanism to convert the extraction task into a sequence labeling task without generating redundant information, but is unable to handle the overlapping relations.

To solve the information redundancy problem and overlapping relation problem described above, we propose a joint model that can handle the sub-tasks of named entity recognition (NER), entity pair extraction (EPE), and relationship classification (RC) simultaneously. The NER sub-task uses the pre-trained BERT (Bidirectional Encoder Representations from Transformers) model [13] to generate word vectors, and takes into account the long-distance dependence of entity labels. The EPE sub-task first uses the proposed Encoder-LSTM network to directly extract the multiple sets of related entity pairs from the sample, then identifies the subject entity and the predicate entity in each entity pair. This approach avoids generating the redundant entity pairs in traditional methods, and also works for overlapping relationship. The RC sub-task uses the traditional relation classification method but taking more abundant and reasonable information as its inputs to improve the performance of classification. In order to solve the problem of information loss between sub-modules and strengthen the interaction between sub-tasks, we designed and added the Connect&LayerNorm layer between sub-modules. We conducted experiments on the public datasets ADE and CoNLL04 to evaluate the effectiveness of our model. The results show that the proposed model achieves good results, and at the same time the model can greatly reduce the amount of redundant information. Compared with other methods, our NER sub-module and RC sub-module have achieved excellent performance. Compared with the traditional LSMT network, the proposed Encoder-LSTM network achieves a significant improvement in performance.

The remainder of the paper is structured as follows. In section 2, we review the related work of named entity recognition, relation classification, and joint extraction tasks. In section 3, we introduce the joint entity and relation extraction model we proposed in detail. In section 4, we first describe the detailed information about the experimental setup, then introduce the experimental results, and analyze the redundancy problem and overlapping relations in detail. Finally, we give the conclusions in Section 5.

## 2. RELATED WORKS

### 2.1. Named Entity Recognition

As a basic task in the field of NLP, NER is to identify the named entities. At present, NER has matured in several research directions. Statistical machine learning-based methods [14], [15], [16] require feature engineering and rely more on corpora. Deep learning-based methods [2], [17], [18] can learn more complex features because of their excellent learning ability. Such methods usually use CNN or RNN to learn sentence features, and then use methods such as conditional random files (CRF) to decode the dependencies between labels, and finally identify the entity label of each token. Deep learning-based methods have also been tried to combine with pre-trained language models such as BERT and achieved excellent performance [19].

### 2.2. Relation Classification

The RC task is a hot research direction in the information extraction task, and its purpose is to determine the category of relationship between two entities in the sentence. Traditional RC methods [20] have good performance on corpora in specific fields, but they rely too much on NLP tools and require a lot of time to design and extract effective features. Due to the advantages of easy learning of complex features, methods based on deep learning [21], [22], [23], [24], [25] have also been widely studied and used by researchers. This type of method uses the original sentence information and the information indicating the entity as inputs to a CNN or RNN to learn the features of a sentence, and finally classifies the constructed relation vector. In recent years, methods based on the combination of deep learning and attention mechanisms have gained significant improvement in performance [26], [27].

### 2.3. Joint Entity and Relation Extraction

The original intention of the method based on joint learning is to overcome the shortcomings of the pipeline-based method. In the early research, feature-based systems [28] can handle two sub-tasks at the same time, but they rely heavily on the features generated by NLP tools and have the problem of propagation errors. To overcome the problems, some methods based on deep learning have been proposed. In 2016, Gupta *et al.* [9] proposed a Table Filling Multi-Task Recurrent Neural Network (TF-MTRNN) model which simplifies the NER and RC tasks into the Table Filling. In 2017, Zheng *et al.* [3] improved the work of [2], proposed a joint model that does not use NLP tools, and solved the problem of long-distance dependence of entity labels. In 2017, Zheng *et al.* [12] proposed a novel labeling mechanism that converts entity and relation extraction task into a sequence labeling task. This method does not generate redundant information. In 2018, to solve the problem of overlapping relations, Bekoulis *et al.* [5] proposed an end-to-end joint model, which treats the extraction task as a multi-head selection problem, so that each entity can judge the relation with other entities. In 2019, Eberts *et al.* [11] proposed a span-based model that achieves the SOTA performance in the field of joint extraction of entity and relation. This model abandons the traditional BIO/BIOU annotation method and consists of three parts: span classification, spam filtering, and relation classification.

Based on the above research, we propose a joint extraction method for information redundancy elimination. Compared with feature-based methods, this method does not require any additional manual features and NLP tools. Compared with previous methods based on deep learning, our method avoid generating redundant information and can handle the overlapping relations.

## 3. MODEL

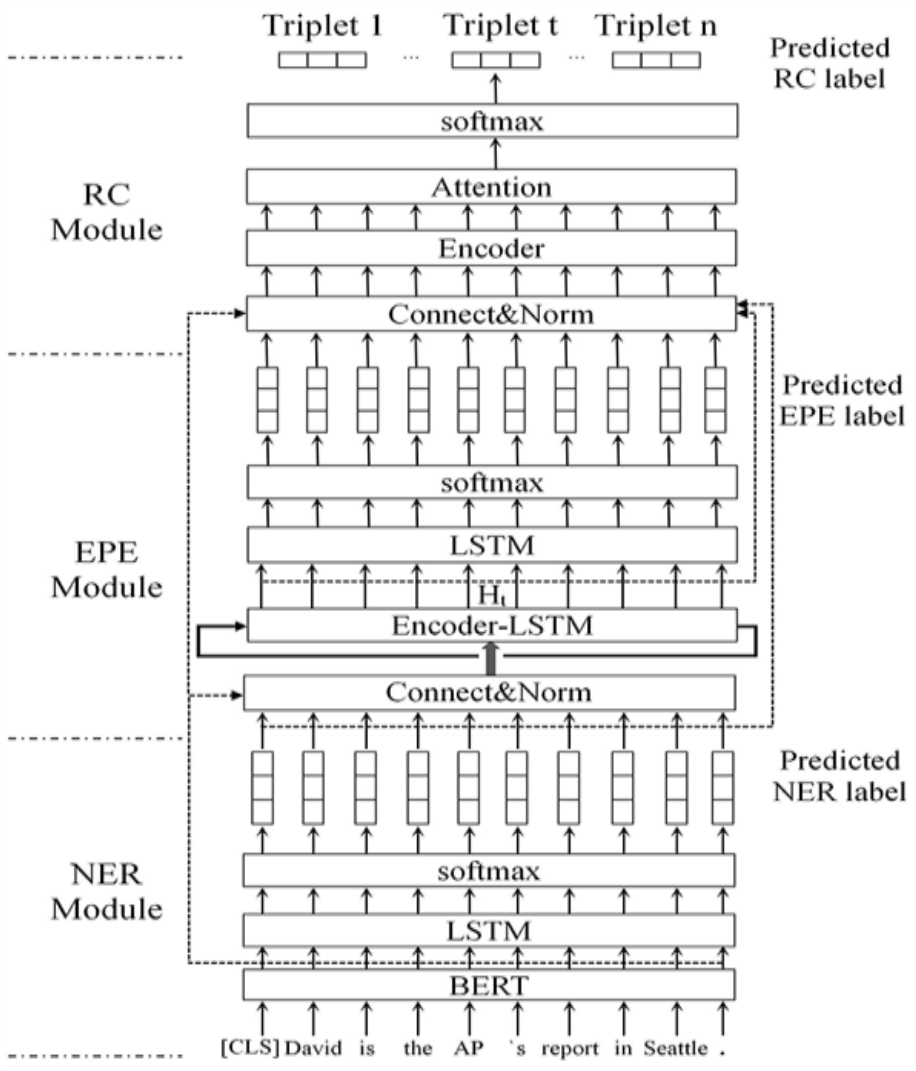

Figure 1. The framework of joint extraction model.

The joint model we proposed consists of three modules: NER module, EPE module, and RC module, as shown in Fig. 1. The NER module identifies the entity label of each token in the text. The EPE module takes sentences and entity labels as inputs, to extracts multiple related entity pairs, and identifies the subject entity and predicate entity for each pair of entities. The RC module classifies the relations.

### 3.1. Named Entity Recognition

The essence of the NER task is sequence labeling, which assigns a label to each token in the sentence. As shown in Fig. 1, the NER module of the proposed model includes a pre-trained BERT model for generating word vectors, an LSTM decoding layer for solving label dependencies [3], and a softmax layer. At first, the NER module inputs the constructed input vector to the BERT model [13] and obtains the word vector of the sentence. The set of word vectors can be expressed as $S=\{w_1,\cdots,w_t,w_{t+1},\cdots,w_l\}\in R^{l\times d}$ , where $w_t$ is the d-dimensional word vector of the t-th word and $l$ is the fixed length of samples. Next, $S$ is inputted to the LSTM decoding layer to perform the following calculation:

$$y_t = LSTM(w_t) \tag{1}$$

where $y_t \in \mathbf{R}^d$, the output of the t-th unit of the decoding layer. Finally the predicted probability of each label of each token of the sentence is obtained through the softmax layer. The predicted probability is expressed as $N=\{p_1,\cdots,p_t,p_{t+1},\cdots,p_l\}\in R^{l\times n_t}$ , where $n_t$ stands for the number of entity labels in the NER module. The loss function of a single sample of this module can be expressed as:

$$L_{ner} = -\sum_{j=1}^{l}\sum_{i=1}^{n_t} Y_{ji}\cdot\log(N_{ji}) \tag{2}$$

where $Y\in R^{l\times n_t}$ is the label of a single sample in the NER module.

Considering the correlation between sub-tasks, we use the original sentence information and the prediction information of the label as the input of the EPE module, denoted as $Z_connect=\{z_1,\cdots,z_t,z_{t+1},\cdots,z_l\}\in R^{l\times(n_t+d)}$ , where $z_t=[w_t;p_t]$ . In addition, we perform LayerNrom [29] processing on the combined input, which is expressed as:

$$Z = LayerNorm(Z_connect) \tag{3}$$

#### 3.2. Entity Pair Extraction

The EPE task is designed to extract multiple pairs of related entities from the inputted sentence. As shown in Fig. 1, the EPE module consists of an Encoder-LSTM network, an LSTM decoding layer, and a softmax layer. Retrieving the pairs of related entities from the sample in a specific order can get a unique sequence, in the form of [(subject entity, predicate entity), ... , (subject entity, predicate entity)]. When the search order is from left to right, the sequence corresponding to the input sample of Fig. 1 takes the form of [(David, AP), (AP, Seattle)]. The order of the sequence is not dependent on whether or not there are overlapping relations among the entities. It is easy to find that the current element pays more attention to the information of the previous element, so we need to retain more new information in each recurrent unit. The addition of new memory in GRU is limited by the old memory, and the update gate in LSTM independently controls how much information in added to the new memory, and the LSTM network can alleviate the problem of gradient disappearance in the traditional RNN model with the long sequence.

Based on the above analysis, the EPE module first uses the Encoder-LSTM network to model the order of the sequence. The output of each recurrent unit of the Encoder-LSTM network is a sentence encoding that contains a pair of related entities. Our proposed Encoder-LSTM network consists of the Encoder structure in Transformer and the LSTM network. The design purpose of the network is to use the Encoder to improve the ability of the recurrent unit to model sentences. The design idea of the network is similar to ConvLSTM [30]. The structure of the Encoder-LSTM network is shown in Fig. 2.

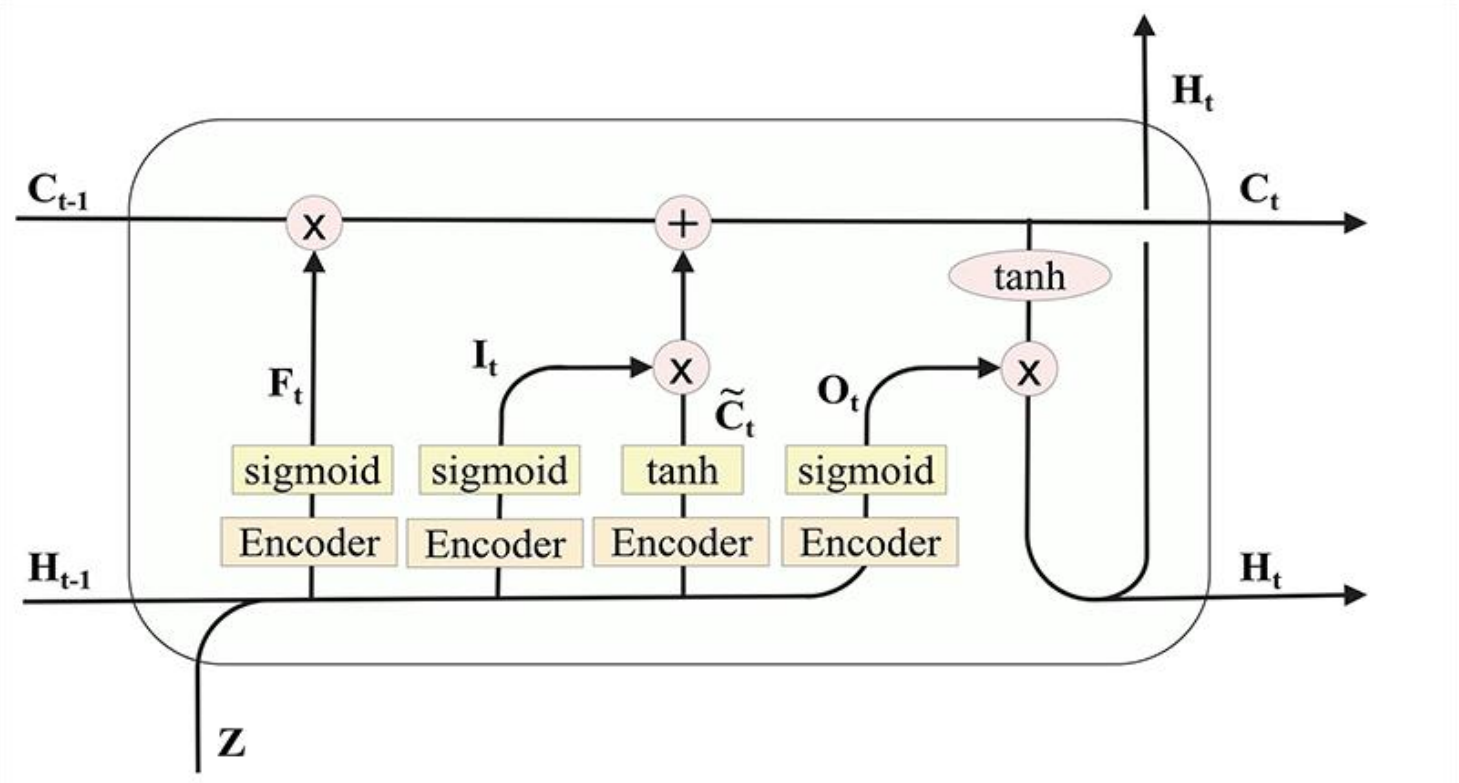

Figure 2. The structure of Encoder-LSTM network.

The network is composed of four parts: the input gate $I_t$, the forget gate $F_t$, the output gate $O_t$, and $\tilde{C}_t$, each of which has its own sentence coding structure $Encoder_i$, $Encoder_f$, $Encoder_o$, and $Encoder_c$ respectively. The calculation details of the Encoder-LSTM network are as follows:

$$I_t = \sigma(Encoder_i([Z;H_{t-1}]) \cdot W_i + b_i) \tag{4}$$

$$F_t = \sigma(Encoder_f([Z;H_{t-1}]) \cdot W_f + b_f) \tag{5}$$

$$O_t = \sigma(Encoder_o([Z;H_{t-1}]) \cdot W_o + b_o) \tag{6}$$

$$\tilde{C}_t = \tanh(Encoder_c([Z;H_{t-1}]) \cdot W_c + b_c) \tag{7}$$

$$C_t = I_t * \tilde{C}_t + C_{t-1} * F_t \tag{8}$$

$$H_t = O_t * \tanh(C_t) \tag{9}$$

where $t = 0,1,\ldots,n$, $n$ stands for the number of related entity pairs being extracted, which is the hyperparameter of the model. $C_t$ and $H_t$ are the state and output of the current recurrent unit respectively. $C_{t-1}$ and $H_{t-1}$ are the state and output before current unit respectively.

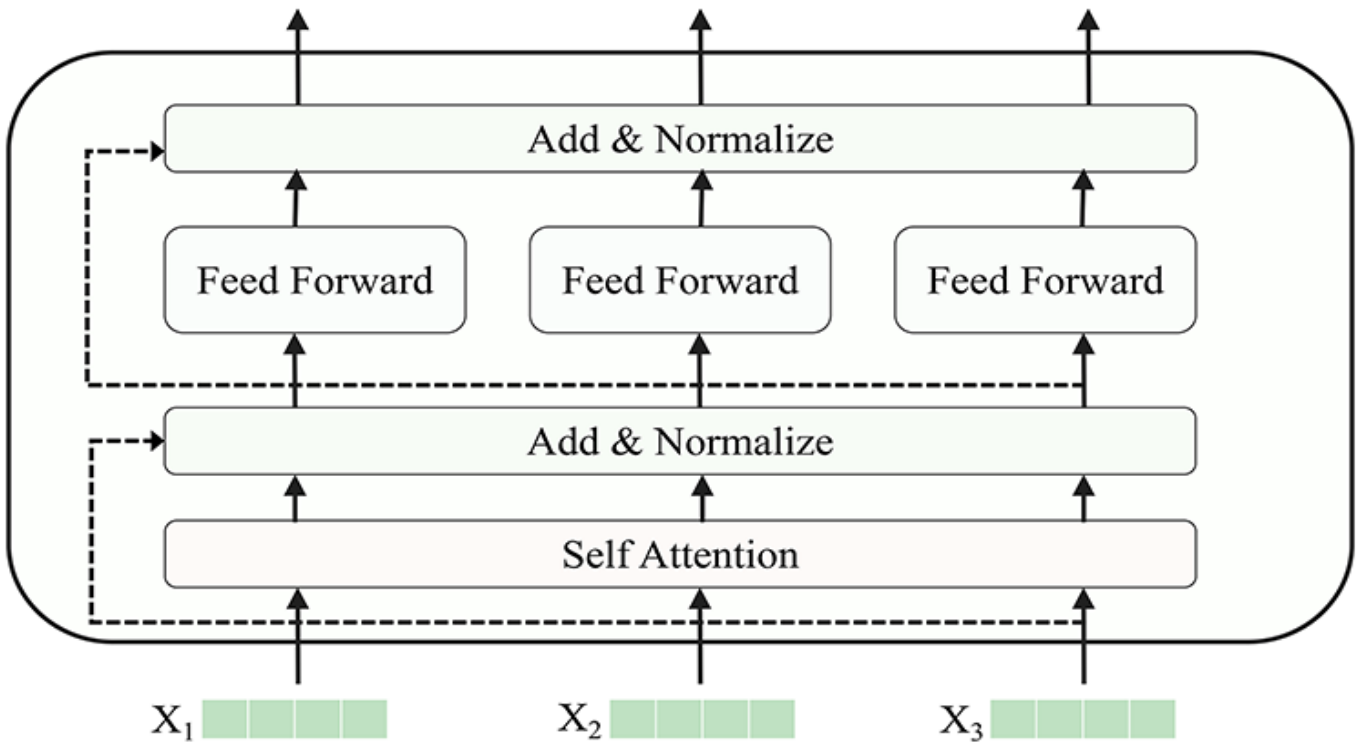

Figure 3. The structure of Encoder.

The function $Encoder_*$ represents the Encoder structure in the Transformer model [29]. The structure of Encoder is shown in Fig. 3. The output of the Encoder-LSTM network is $n$ sets of sentence encoding $H = \{H_1, \ldots, H_t, \ldots, H_n\} \in R^{n \times l \times d_w}$, where $d_w$ is the dimension of the hidden layers of the networks.

The relation type of the entity pair is determined by both the types of the subject entity and the predicate entity. Just knowing the categories of two entities is not sufficient to determine the relationship of the entity pair. Therefore, the EPE module should be able to identify the subject-predicate label of entities in the sentence encoding. The EPE module takes $H$ as input and predicts the subject-predicate label of entities through the LSTM decoding layer and the softmax layer. The prediction probability of the subject-predicate label is expressed as $M = \{M_1, M_2, \cdots, M_n\} \in R^{n \times l \times n_d}$. The loss function of a single sample of this module can be expressed as:

$$L_{epe} = -\sum_{k=1}^{n}\sum_{j=1}^{l}\sum_{i=1}^{n_d} Y_{kji} \cdot \log(M_{kji}) \tag{10}$$

where $Y \in R^{n \times l \times n_d}$ is the subject-predicate label of a single sample, and $n_d$ is the number of subject-predicate labels in the EPE module.

### 3.3. Relation Classification

The goal of the RC module is to classify the relations of entity pairs that have been specified by subject-predicate labels. As shown in Fig. 1, this module consists of Encoder structure, Attention mechanism, and softmax layer.

The input of the traditional RC task contains not only sentence encoding information but also position information indicating two entities. This is different from the RC task of the previous joint method that only uses inter-entity sentence information [3] or two tokens as input information [5]. To improve the performance of the RC task, we adopt the idea of Position Feature [23] and Position Indicators [24], and use the predicted subject-predicate labels $M_t$ of entities as the position indicator of two entities. In addition, in order to strengthen the interaction between sub-tasks and solve the problem of information loss between sub-tasks , the input of the RC task also includes the information of NER module. Finally, the RC task takes the concatenation of the sentence encoding $H_t$ , the predicted subject-predicate label $M_t$ , the predicted entity label $N$, and the word vectors $S$ as the input, which can be expressed as $\left[M_t;H_t;N;S\right]\in R^{l\times(d_w+n_d+d+n_t)}, t\in 1,2,\ldots n$. Next we perform LayerNorm [29] processing on the input.

$$LN_t = LayerNorm(\left[M_t;H_t;N;S\right]) \tag{11}$$

To improve the performance, the RC module first uses the Encoder structure to learn sentence features.

$$L_t = Encoder_r(LN_t) \tag{12}$$

then the features are processed by the Attention mechanism [31] to get the relation vector.

$$r_t = \text{Attention}(L_t) \tag{13}$$

where $r_t\in\mathbf{R}^{d_w+n_d}$ represents the relation vector of the t-th related entity pair. Finally, the module obtains the predicted probability $P=\{p_1,\cdots,p_t,\cdots,p_n\}\in R^{n\times n_r}$ of the relation category through the softmax layer, where $p_t\in\mathbf{R}^{n_r}$ is the prediction probability of the relation category of the t-th entity pair, and $n_r$ is the number of relation categories. The loss function of a single sample of this module can be expressed as:

$$L_{rc} = -\sum_{j=1}^{n}\sum_{i=1}^{n_r} Y_{ji}\cdot\log(P_{ji}) \tag{14}$$

where $Y\in R^{n\times n_r}$ is the relation label of a single sample.
Different from the traditional joint model, our model performs the task of entity and relation extraction with three sub-modules, and the final loss is the sum of the three parts: $L_{all} = L_{ner} + L_{epe} + L_{rc}$.

# 4. EXPERIMENT AND ANALYSIS

## 4.1. Experimental Setting

DATASET: We conducted experiments on two datasets: (i) Adverse Drug Events, ADE dataset [32]， and (ii) the CoNLL04 dataset [33]. ADE: The dataset includes two entity types Drug and Adverse-Effect and a single relation type Adverse-Effect. There are 4272 sentences and 6821 relations in total and similar to previous work [11], we remove ~120 relations with overlapping entities. Since there are no official test set, we evaluate our model using 10-fold cross-validation similar to previous work [11]. The final results are displayed in F1 metric as a macro-average across the folds. We adopt strict evaluation setting to compare to previous work [5], [11], [34], [35]. CoNLL04: The dataset contains four entity types (Location, Organization, Person, Other) and five relation types (Kill, Live_in, Located_in, OrgBased_in, Work_for). For the dividing rules of the dataset, the experiment follows the method defined by Gupta *et al.* [9]. The original 1441 samples are divided into the training set, the validation set, and the test set, with 910, 243, and 288 samples respectively. We adopt relaxed evaluation setting to compare to previous work [5], [9], [10]. We measure the performance by computing the average F1 score on the test set.

BASELINES: The baselines we used are recent methods for the ADE dataset and the CoNLL04 dataset. Method Li *et al.* (2016) [34] and method Li *et al.* (2017) [35] have achieved good results on the ADE corpus using a joint model based on parameter sharing. Methods Gupta *et al.*(2016)

[9] and Adel&Schütze(2017) [10] formulate joint entity and relation extraction as a table-filling problem. Method Bekoulis *et al.* (2018) [5] employ a bidirectional LSTM to encode words and use a sigmoid layer to output the probability of a specific relation between two words that belong to an entity. Method Eberts *et al.* (2019) [11]proposed a span-based joint model that relies on the pre-trained Transformer network BERT as its core and achieves the best results.

METRICS: To compare with the previous research, the experiment will evaluate the performance of the three sub-tasks by the values of Precision, Recall, and F1-measure. We use two different settings to evaluate performance, namely strict and relaxed. In the strict setting, an entity is considered correct if the boundaries and the type of the entity are both correct; an entity pair is considered correct if the boundaries and the type of the subject entity and the predicate entity are both correct and the argument entities are both correct; a relation is correct when the type of the relation and the argument entity pair are both correct. In the relaxed setting, the experiment will assume that the boundary of the entities is known, an entity is considered correct if the type of any token of the entity is correctly classified; an entity pair is correct when the type of any token of the subject entity and the predicate entity are both correct and the argument entities are both correct; a relation is correct when the type of the relation and the argument entity pair are both correct. The formulas for Precision, Recall, and F1 are as follows.

$$Precision = \frac{TP}{TP + FP} \tag{15}$$

$$Recall = \frac{TP}{TP + FN} \tag{16}$$

$$F_1\text{-}measure = \frac{2 \times Precision \times Recall}{Precision + Recall} \tag{17}$$

HYPERPARAMETERS: The experiment uses the language Python, the TensorFlow libraries, and the pretrained BERT model of cased_L-12_H-768_A-12 to implement the joint model. For our training on the ADE dataset, the learning rate, the batch size, and the number of iterations are 0.00002, 8, and 40 respectively. The fixed length of the sentence is 128. The value of Dropout is varied for modules and ranging from 0.3 to 0.5. The number of hidden layer units in the Encoder-LSTM network is 96, and the hyperparameter $n$ is 3. The number of layers, the number of heads in Encoder-LSTM network are 2, 4 respectively. We adjusted the hyperparameters of the model for different datasets. The experiment was conducted on an Nvidia DGX-1 server equipped with 8 TeslaV100 GPUs with 128GB of memory per GPU.

## 4.2. Results

The final experimental results are shown in Table 1. The first column indicates the considered dataset. The second column is the comparable previous methods and ours. The results of the NER task (Precision, Recall, F1) are shown in the next three columns, then follows the results of EPE and RC task. Since the EPE task is proposed for the first time in this paper, there are no comparable results for this task. The last column gives the average F1 of all sub-tasks (Overall F1).

Table 1. Comparisons with the different methods.

| Dataset | Methods | NER | | | EPE | | | RC | | | Overall F1 |
|---|---|---|---|---|---|---|---|---|---|---|---|
| | | P | R | $F_1$ | P | R | $F_1$ | P | R | $F_1$ | |
| ADE | Li *et al.* (2016) | 79.50 | 76.60 | 79.50 | - | - | - | 64.00 | 62.90 | 63.40 | 71.45 |
| | Li *et al.* (2017) | 82.70 | 86.70 | 84.60 | - | - | - | 67.50 | 75.80 | 71.40 | 78.00 |
| | Bekoulis *et al.* (2018) | 84.72 | 88.16 | 86.40 | - | - | - | 72.10 | 77.24 | 74.58 | 80.49 |
| | Eberts *et al.* (2019) | 89.26 | 89.26 | 89.25 | - | - | - | 77.77 | 79.96 | 78.84 | 84.05 |
| | Proposed(Encoder-LSTM) | 90.54 | 92.69 | 91.60 | 80.17 | 80.03 | 80.10 | 77.63 | 80.03 | 78.81 | 83.50 |
| CoNLL04 | Gupta *et al.* (2016) | 88.50 | 88.90 | 88.80 | - | - | - | 64.40 | 53.10 | 58.30 | 73.60 |
| | Adel&Schütze (2017) | - | - | 82.10 | - | - | - | - | - | 62.50 | 72.30 |
| | Bekoulis *et al.* (2018) | 93.41 | 93.15 | 93.26 | - | - | - | 72.99 | 63.37 | 67.01 | 80.14 |
| | Proposed(Encoder-LSTM) | 91.87 | 96.45 | 94.11 | 68.42 | 67.16 | 67.78 | 66.42 | 63.23 | 64.79 | 75.56 |

For the ADE dataset, we can observe that in the NER task, the Proposed(Encoder-LSTM) method achieves the best performance. The macro-F1 value of this method is 2.5% higher than that of the Eberts *et al.* (2019) method. In the EPE task, the macro-F1 value of the Proposed(Encoder-LSTM) method is 83%. In the RC task, the Proposed (Encoder-LSTM) method has significantly improved the macro-F1 value compared to the Li *et al.* (2016) method, Li *et al.* (2017) method, and Bekoulis *et al.* (2018) method, and has similar performance compared to the Eberts *et al.* (2019) method.

Considering the results in the CoNLL04 dataset, we can observe that the Proposed(Encoder-LSTM) method achieves the best results in the NER task. Compared with method Bekoulis *et al.* (2018), the Proposed(Encoder-LSTM) method has a significant improvement in F1 value. In the EPE task, the F1 value of the Proposed(Encoder-LSTM) method is 67.78%. In the RC task, the Proposed(Encoder-LSTM) method achieves good results. Compared with method Adel&Schü tze(2017), the F1 value of the Proposed(Encoder-LSTM) method is increased by about 2.3%.

It can been seen from the results that our model has achieved excellent performance on both NER and RC modules, but the overall performance of our model is similar to the comparison methods. The reason for the above phenomenon is that the performance of EPE module has become the bottleneck of the overall performance of the model. It can be noticed that there are differences in the performance of the model on the two datasets. After analysis, this is related to the number of samples containing multiple related entity pairs in the dataset. Because our model extracts entity pairs by learning the order relationship of related entity pairs, the ADE dataset can provide more effective data than the CoNLL04 dataset.

Table 2. Ablation tests on the ADE dataset.

| Settings | NER F1(%) | EPE F1(%) | RC F1(%) | Overall F1(%) |
|---|---|---|---|---|
| Proposed | 91.59 | 80.09 | 78.91 | 83.50 |
| - LSTM Decoder | 91.32 | 79.73 | 78.81 | 83.28 |
| - Connect&LayerNorm | 91.56 | 78.51 | 76.66 | 82.24 |
| - Encoder-LSTM | 91.04 | 76.24 | 74.81 | 80.69 |

We conduct ablation tests on the ADE dataset reported in Table 2 to analyze the effectiveness of the Encoder-LSTM network and other components in the model. The performance of the model decreases (~0.2% in terms of Overall F1 score) when we remove the LSTM decoder layer. This shows that the LSTM Decoder layer can strengthen the ability of model to learn the dependency between entity tags [3]. The performance of EPE and RC tasks decreases (~1.2%) when we remove the Connect&LayerNorm layer of the RC module and only use the predicted subject-predicate labels and the sentence encoding as inputs for the RC task. This shows that the predicted entity labels and the word vectors provide meaningful information for the RC component and this approach can solve the problem of information loss between subtasks. There is also a reasonable explanation that this approach is similar to the residual structure [29], which can alleviate the problem of gradient disappearance. Finally we conduct experiments by removing the Encoder-LSTM network and substituting it with a LSTM network. This approach leads to a slight decrease in the F1 performance of the NER module, while the performance of the EPE task and the RC task decreased by about 2%. This happens because the Encoder structure in the Encoder-LSTM network can improve the ability of recurrent units to model sentences.

Table 3. Model performance for different hyperparameter values.

| Hyper-parameters | value | NER F1(%) | EPE F1(%) | RC F1(%) | Overall F1(%) |
|---|---|---|---|---|---|
| Encoder-layer | 2 | 91.46 | 78.86 | 77.20 | 82.51 |
| | 3 | 91.59 | 80.09 | 78.81 | 83.50 |
| | 4 | 91.59 | 78.62 | 77.36 | 82.52 |
| hidden size | 32 | 91.25 | 77.58 | 75.75 | 81.52 |
| | 64 | 90.73 | 78.42 | 77.17 | 82.10 |
| | 96 | 91.59 | 80.09 | 78.81 | 83.50 |
| | 128 | 91.60 | 78.96 | 77.43 | 82.66 |

We also evaluated the impact of different hyperparameter values in the Encoder-LSTM network on model performance. Table 3 show the performance of our model on the ADE dataset for different values of Encoder layer and hidden size hyperparameters in Encoder-LSTM network, respectively. It can be observed that the model achieves the best performance with the encoder layers of 3 and the hidden size of 96.

## 4.3. Analysis of Redundancy and Overlapping Relation

The redundancy problem means that the model generates and has to evaluate a large number of unrelated entity pairs. The method we proposed directly extracts the pairs of related entities from the samples, without producing redundant information in the traditional sense. In order to solve the problem of different numbers of triples in different samples, our method uses the hyperparameter $n$ to specify the number of related entity pairs extracted in each sample, but this approach leads to the inevitable generation of redundant sentence coding in the EPE module.

Because the redundancy of the model is proportional to the number of times the model classifies the relationships, we use this number to evaluate and compare the redundancy of different models. The method proposed by Miwa *et al.* [8] labels m(m-1)/2 cells in the entity and relation table to predict possible relationships, where m is the sentence length. The method by Zheng *et al.* [3] and Bekoulis *et al.* method [5] first identify entities, and then classify the relationships between each pair of entities, so these two methods classify the relationships $k^2$ times, where k is the number of identified entities. Our method directly extracts the related entity pairs and then classifies the relationships of each entity pair. Therefore, the number of times our method classifies the relationships is equal to the number $n$ of related entity pairs extracted by the model, and $n$ is the hyperparameter of our model. Based on the above analysis, we obtain a statistical table of the number of times the model classifies the relationships, as shown in Table 4.

Table 4. Redundancy of different models.

| Methods | times |
|---|---|
| Miwa&Sasaki(2014) | $m(m-1)/2$ |
| Zheng *et al.*(2017) | $k^2$ |
| Bekoulis *et al.*(2018) | $k^2$ |
| Proposed | $n$ |

The parameter m, k and $n$ in the Table 4 stand for the sentence length, the number of entities, and the hyperparameter of our model respectively.

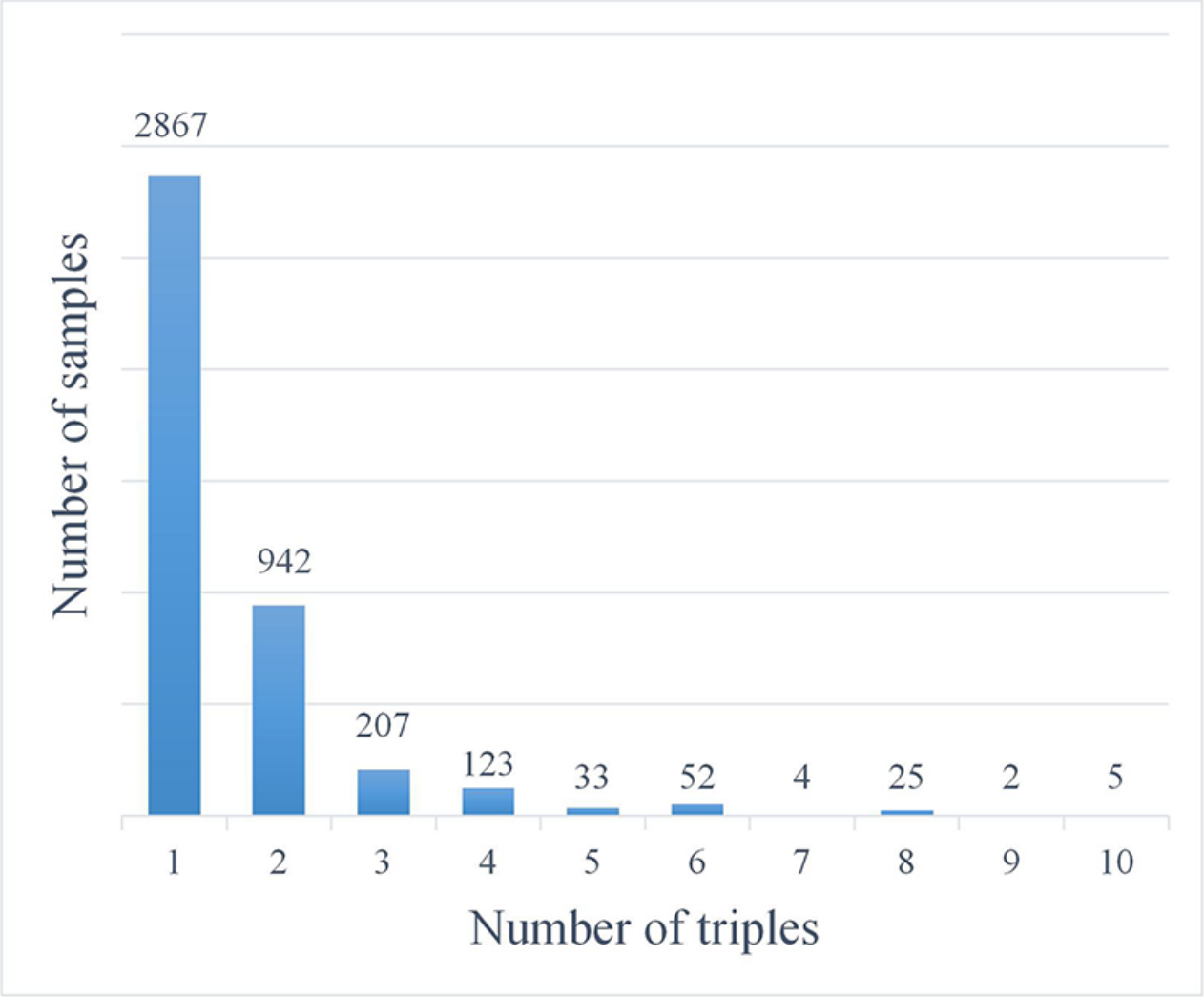

Figure 4. Distribution histogram of the number of triples in the ADE dataset.

After analysis, more than 99% of the word pairs are irrelevant [9]. About 45% of the samples contain more than 3 entities, and the related entity pairs only account for a small part of all entity pairs. As shown in Fig. 4, about 77% of the samples contain only one triple, and about 96% of the samples contain no more than three triples. For example, assuming the input sample is shown in Fig. 1, then m, k, and $n$ take the value of 128, 3, and 3 respectively. The number of times of Miwa&Sasaki(2014) method, Zheng *et al.* (2017) method, Bekoulis *et al.* (2018) method, and our method are 8128, 9, 9, and 3 respectively. Therefore, if the value of $n$ is appropriately selected, the redundancy of the proposed method is much smaller than that of other methods.

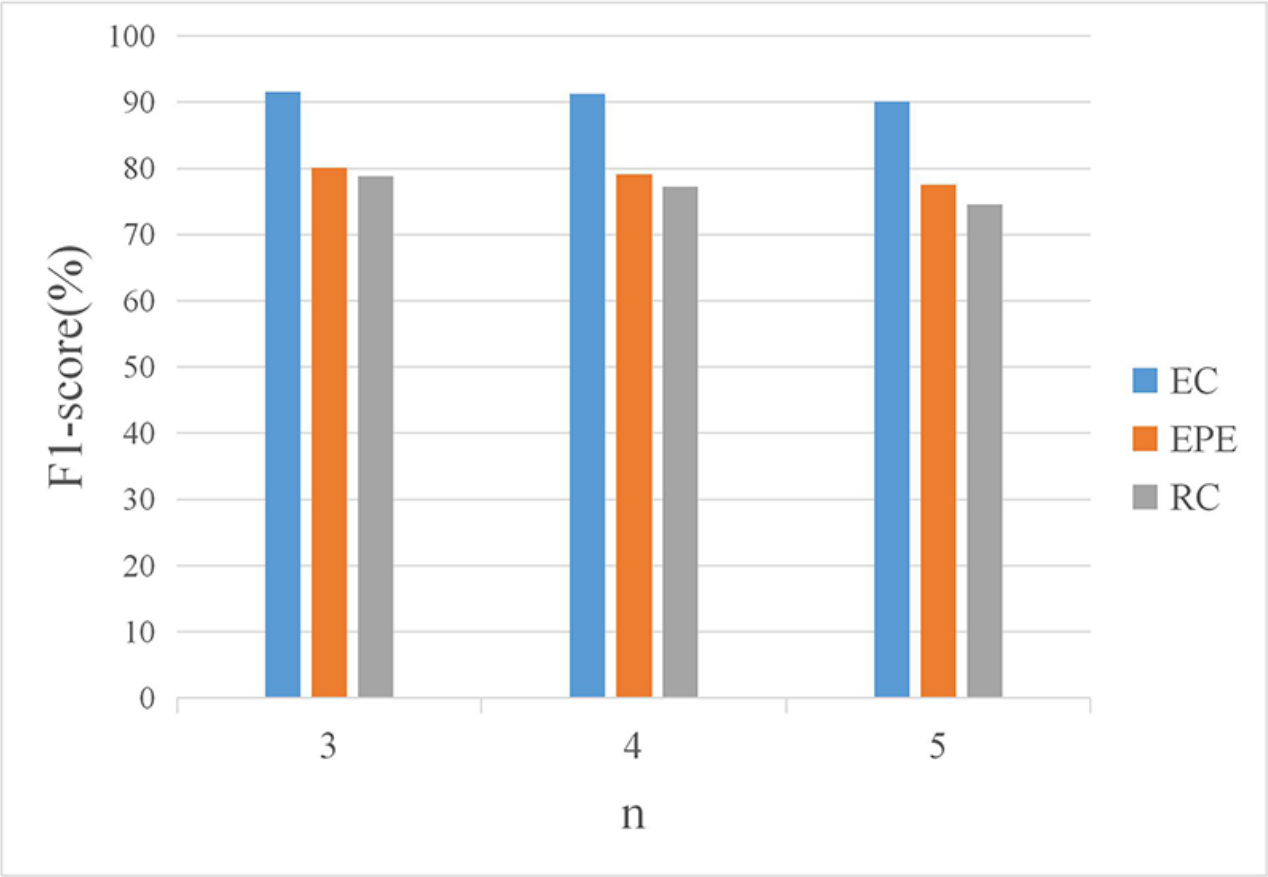

Figure 5. Model performance for different values hyper-parameter n.

Since the redundancy of our model depends on the value of $n$, to evaluate the impact of redundancy on performance, we conduct experiments based on different values of $n$, and the results are shown in Fig. 5. It can be observed that the model has the best overall performance when the hyperparameter $n$ is 3. The change of the value of $n$ has little effect on the performance of the NER module and the EPE module. As the value of $n$ increases, the performance of the RC module and the EPE module decreases significantly. After analysis, this phenomenon is related to the distribution of the number of triples in the sample. Theoretically, as the value of $n$ increases, the EPE module can better model the sequence information of related entity pairs. However, it can be seen from Fig. 4 that there are very few useful data when $n$ is greater than 3. At this time, the increase of the value of $n$ not only cannot help the learning of the EPE module, but also seriously interferes with the training of the model. Based on the above analysis, the choice of $n$ value should depend on the distribution of the number of triples in the sample. If the samples in the corpus contain sufficient related entity pairs, our model will perform better, otherwise our model will perform not well.

There are two types of overlapping relations [36]. The first type is that an entity has relations with multiple other entities. Our EPE module uses the order information of the sequence of related entity pairs to extract entity pairs. This type of overlapping relations does not affect the unique order of the sequence. Therefore, the proposed method works well with such situation. The second type of overlapping relations refers to the multiple relationships between one entity pair. Since this situation does not exist in the ADE dataset and the CoNLL04 dataset, we treat the RC task as a multiclass classification task to evaluate which relationship category the entity pair belongs to. Specifically, our model uses the softmax function as the activation function of the output layer, and the categorical cross-entropy as the loss function. If we need to deal with the second kind of overlapping relations, we can treat the RC task as a multilabel classification task, such as the Bekoulis method [5], to evaluate the various relationships that may exist in the entity pair. Specifically, our model uses the sigmoid function as the activation function of the output layer, and uses binary cross-entropy as the loss function.

## 5. CONCLUSION

We have presented the joint extraction model based on entity pair extraction with information redundancy elimination. The model first extracts multiple sets of sentence encoding from the sample, then identifies the subject entity and the predicate entity in each set of sentence encoding, and finally classifies the relationship between the two entities. We also propose the Encoder-LSTM network, which improves the ability of recurrent units to model sentences. By conducting experiments on the ADE dataset and the CoNLL04 dataset, we verified the effectiveness of the method and evaluated the performance of the model. Compared with other joint extraction methods, our method solves the problem of redundancy of unrelated entity pairs while achieving excellent performance, and can handle the cases with overlapping relationships.

Since the performance of our EPE module limits the overall model, as the future work we will try to optimize the solution of the EPE. And we plan to verify the proposed method on more actual datasets.

## ACKNOWLEDGEMENTS

The work is partially supported by the Shaanxi Key Laboratory of Network Data Analysis and Intelligence Processing. Our deepest gratitude also goes to the anonymous reviewers for their suggestions for improving this paper.

## REFERENCES

[1] J. Li, A. Sun, J. Han, C. Li "A Survey on Deep Learning for Named Entity Recognition," IEEE *Transactions on Knowledge and Data Engineering*, vol. pp, no. 99, pp. 1–1, Mar. 2020.

[2] M. Miwa and M. Bansal, "End-to-end relation extraction using LSTMs on sequences and tree structures," in *Proc. Annu. Meet. Assoc. Comput. Linguist.*, Berlin, Germany, 2016, pp. 1105–1116.

[3] S. Zheng, Y. Hao, D. Lu, H. Bao, J. Xu, H. Hao, and B. Xu, "Joint entity and relation extraction based on a hybrid neural network," *Neurocomputing*, vol. 257, pp. 59–66, Sep. 2017.

[4] Z. Peng, S. Zheng, J. Xu, Z. Qi, and X. Bo, "Joint Extraction of Multiple Relations and Entities by Using a Hybrid Neural Network," in *Proc. Lect. Notes Comput. Sci.*, Nanjing, China, 2017, pp. 135–146.

[5] G. Bekoulis, J. Deleu, T. Demeester, and C. Develder, "Joint entity recognition and relation extraction as a multi-head selection problem," *Expert Systems with Applications*, vol. 114, pp. 34–45, Dec. 2018.

[6] G. Bekoulis, J. Deleu, T. Demeester, and C. Develder, "An attentive neural architecture for joint segmentation and parsing and its application to real estate ads," *Expert Systems with Applications*, vol. 102, pp. 100–112, Jul. 2018.

[7] X. Zhang, J. Cheng, and M. Lapata, "Dependency parsing as head selection," in *Proc. Conf. Eur. Chapter Assoc. Comput. Linguist.*, Valencia, Spain, 2017, pp. 665–676.

[8] M. Miwa and Y. Sasaki, "Modeling joint entity and relation extraction with table representation," in *Proc. Conf. Empir. Methods Nat. Lang. Process.*, Doha, Qatar, 2014, pp. 1858–1869.

[9] P. Gupta, H. Schütze, and B. Andrassy, "Table filling multi-task recurrent neural network for joint entity and relation extraction," in *Proc. Int. Conf. Comput. Linguist.*, Osaka, Japan, 2016, pp. 2537–2547.

[10] H. Adel and H. Schütze, "Global normalization of convolutional neural networks for joint entity and relation classification," in *Proc. Conf. Empir. Methods Nat. Lang. Process.*, Copenhagen, Denmark, 2017, pp. 1723–1729.

[11] M. Eberts, A. Ulges, "Span-based Joint Entity and Relation Extraction with Transformer Pre-training," *arXiv*, 2019.

[12] S. Zheng, F. Wang, H. Bao, Y. Hao, P. Zhou, and B. Xu, "Joint extraction of entities and relations based on a novel tagging scheme," in *Proc. Annu. Meet. Assoc. Comput. Linguist.*, Vancouver, Canada, 2017, pp. 1227–1236.

[13] J. Devlin, M. Chang, K. Lee, and K. Toutanova, "BERT: pre-training of deep bidirectional transformers for language understanding," *CoRR*, vol. abs/1810.04805, 2018.

[14] F. Alam and I. Asiful, , "A proposed model for Bengali named entity recognition using maximum entropy markov model incorporated with rich linguistic feature set," in *ACM Int. Conf. Proc. Ser.*, Dhaka, Bangladesh, 2020.

[15] L.Gong, X. Liu, X. Yang, L. Zhang, Y. Jia, and R. Yang, "CBLNER: A Multi-models Biomedical Named Entity Recognition System Based on Machine Learning," in *Lect. Notes Comput. Sci.*, Nanchang, China, 2019, pp. 51–60.

[16] A. Anandika, S. Mishra, "A study on machine learning approaches for named entity recognition," in *Proc. -Int. Conf. Appl. Mach. Learn.*, Bhubaneswar, India, 2019, pp. 153–159.

[17] H. Wei *et al.*, "Named Entity Recognition From Biomedical Texts Using a Fusion Attention-Based BiLSTM-CRF," *IEEE Access*, vol. 7, pp. 73627–73636, June. 2019.

[18] S. Zhang, Y. Shen, J. Gao, J. Chen, J. Huang, and S. Lin, "A Multi-domain Named Entity Recognition Method Based on Part-of-Speech Attention Mechanism," in *Commun. Comput. Info. Sci.*, Kunming, China, 2019, pp. 631–644.

[19] Z. Dai, X.Wang, P. Ni, Y. Li, G. Li, and X. Bai, "Named entity recognition using bert bilstm crf for chinese electronic health records," in *Proc. Int. Congr. Image Signal Process., BioMed. Eng. Inf.*, Huaqiao, China, 2019, pp. 1–5.

[20] B. Rink and S. Harabagiu, "UTD: Classifying semantic relations by combining lexical and semantic resources," in *Proc. Int.Workshop Semant. Evaluation*, Uppsala, Sweden, 2010, pp. 256–259.

[21] X. Guo, H. Zhang, H. Yang, L. Xu and Z. Ye, "A Single Attention-Based Combination of CNN and RNN for Relation Classification," *IEEE Access*, vol. 7, pp. 12467–12475, 2019.

[22] C. Zhang *et al.*, "Multi-Gram CNN-Based Self-Attention Model for Relation Classification," *IEEE Access*, vol. 7, pp. 5343–5357, 2019.

[23] D. Zeng, K. Liu, S. Lai, G. Zhou, and J. Zhao, "Relation classification via convolutional deep neural network," in *Proc. COLING - Int. Conf. Comput. Linguist.*, Dublin, Ireland, 2014, pp. 2335–2344.

[24] D. Zhang and D. Wang, "Relation classification via recurrent neural network," *CoRR*, vol. abs/1508.01006, 2015.

[25] X. Huang, J. Lin, W. Teng and Y. Bao, "Relation classification via CNNs with Attention Mechanism for Multi-Window-Sized Kernels," in *Proc. IEEE Adv. Inf. Technol., Electron. Autom. Control Conf.,* Chengdu, China, 2019, pp. 62-66.

[26] L. Wu, H. Zhang, H. Yang, Y. Yang, X. Liu, and K. Gao, "Dynamic Prototype Selection by Fusing Attention Mechanism for Few-Shot Relation Classification," in *Lect. Notes Comput. Sci.*, Phuket, Thailand, 2020, pp. 431–441.

[27] S. Zhang, D. Zheng, X. Hu, and M. Yang, "Multi-Channel CNN Based Inner-Attention for Compound Sentence Relation Classification," *IEEE Access*, vol. 7, pp. 141801–141809, 2019.

[28] Q. Li and H. Ji, "Incremental joint extraction of entity mentions and relations," in *Proc. Annu. Meet. Assoc. Comput. Linguist.*, Baltimore, MD, United states, 2014, pp. 402–412.

[29] A. Vaswani, N. Shazeer, N. Parmar, J. Uszkoreit, L. Jones, A. N. Gomez, L. Kaiser, and I. Polosukhin, "Attention is all you need," in *Adv. neural inf. proces. syst.*, Long Beach, CA, United states, 2017, pp. 5999–6009.

[30] X. Shi *et al.*, "Convolutional LSTM Network: A Machine Learning Approach for Precipitation Nowcasting,"in *Conf. Advances in Neural Information Processing Systems*, Montreal, QC, Canada, 2015, pp. 802 - 810.

[31] K. Xu, J. Ba, R. Kiros, K. Cho, A. Courville, R. Salakhutdinov, R. Zemel, and Y. Bengio, "Show, attend and tell: Neural image caption generation with visual attention," in *Proc. Int. Conf. Mach. Learn.*, Lile, France, 2015, pp. 2048–2057.

[32] Gurulingappa *et al.* "Development of a benchmark corpus to support the automatic extraction of drug-related adverse effects from medical case reports,"*Biomedical Informatics*, vol. 45, no. 5, pp. 885–892, 2012.

[33] D. Roth and W. Yih, "A linear programming formulation for global inference in natural language tasks," in *Proceedings of the Eighth Conference on Computational Natural Language Learning*, Boston, MA, USA, 2004, pp. 1–8.

[34] F. Li, Y. Zhang, M. Zhang, and D. Ji, "Joint Models for Extracting Adverse Drug Events from Biomedical Text,"in *Conf. Int. Joint Conf. Artif. Intell.*, New York, NY, United states, 2016, pp. 2838 - 2844.

[35] F. Li, M. Zhang, G. Fu, and D. Ji, "A neural joint model for entity and relation extraction from biomedical text,"*Bmc Bioinformatics*, vol. 18, no. 1, pp. 198, 2017.

[36] S. Wang, Y. Zhang, W. Che, and T. Liu, "Joint extraction of entities and relations based on a novel graph scheme," in *Proc. IJCAI Int. Joint Conf. Artif. Intell.*, Stockholm, Sweden, 2018, pp. 4461–4467.

## AUTHORS

**Yuanhao Shen**

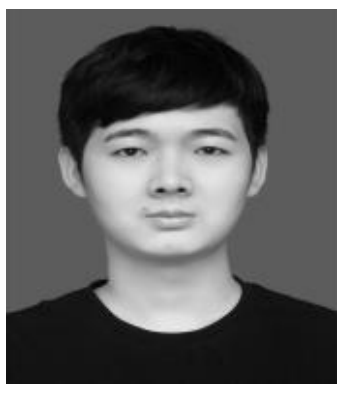

He received the B.E. degree from Xi`an University of Posts and Telecommunications, China, 2018. He is currently pursuing the master's degree in the College of Computer Science and Technology. Hisl research interests include natural language processing and deep learning.

**Jungang Han**

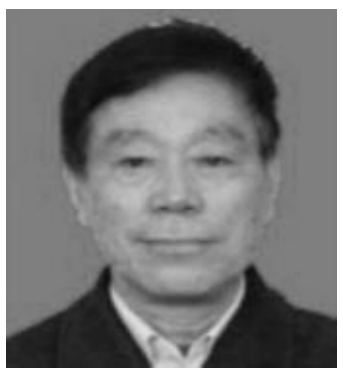

He is a professor at Xi'an University of Posts and Telecommunications. He is the author of two books, and more than100 articles in the field of computer science. His current research interests include artificial intelligence, deep learning for medical image processing.